\documentclass[%
 reprint,
 amsmath,amssymb,
 aps,
]{revtex4-1}

\pdfoutput=1

\usepackage{graphicx}
\usepackage{dcolumn}
\usepackage{bm}

\DeclareMathOperator*{\argmin}{arg\,min}
\DeclareMathOperator*{\argmax}{arg\,max}

\begin{document}

\newcommand{\avg}[1]{\mbox{$\left\langle \, #1 \, \right\rangle$}}
\newcommand{\qav}[1]{\mbox{$\left\langle\left\langle \, #1 \, \right\rangle\right\rangle$}}
\newcommand{\fish}[1]{\mbox{$J\left[ \, #1 \, \right]$}}
\newcommand{\ra}{\rightarrow}
\newcommand{\snr}{\text{SNR}}

\newcommand{\qsopt}{q_s^\text{opt}}
\newcommand{\qeopt}{q_\epsilon^\text{opt}}
\newcommand{\qopt}{q^\text{opt}}
\newcommand{\qoptrho}{q_{d}^{\text{opt}}}

\newcommand{\rhoopt}{\rho^\text{opt}}
\newcommand{\sigopt}{\sigma^\text{opt}}
\newcommand{\pmf}{P_{\text{MF}}}

\newcommand{\sml}{\mathbf{\hat{s}^{\text{ML}}}}
\newcommand{\smap}{\mathbf{{\hat{s}}^{\text{MAP}}}}
\newcommand{\smmse}{{\mathbf{\hat{s}}^{\text{MMSE}}}}

\newcommand{\ze}{z_{\epsilon}}
\newcommand{\zs}{z_s}
\newcommand{\peps}{P_\epsilon}
\newcommand{\pepsq}{P_{\epsilon, q^{\text{opt}}}}
\newcommand{\psrho}{P_{s,\qrho^{\text{opt}}}}

\newcommand{\mor}[3]{\mbox{$\mathcal{M}_{#1}[\, #2 \, ](#3)$}}
\newcommand{\mornoarg}[2]{\mbox{$\mathcal{M}_{#1}[\, #2 \, ]$}}

\newcommand{\morp}[3]{\mbox{$\mathcal{M}'_{#1}[\, #2\, ](#3)$}}
\newcommand{\morpp}[3]{\mbox{$\mathcal{M}''_{#1}[\, #2 \, ](#3)$}}

\newcommand{\proxop}[3]{\mbox{$\mathcal{P}_{#1}[\, #2 \, ](#3)$}}
\newcommand{\proxopnoarg}[2]{\mbox{$\mathcal{P}_{#1}[\, #2 \, ]$}}
\newcommand{\proxopp}[3]{\mbox{$\mathcal{P}'_{#1}[\, #2 \, ](#3)$}}

\newcommand{\lrho}{\lambda_{\rho}}
\newcommand{\lsig}{\lambda_{\sigma}}
\newcommand{\qs}{q_{s}}

\newcommand{\qrho}{q_{d}}
\newcommand{\Egen}{\mathcal{E}^\text{gen}}
\newcommand{\Etrain}{\mathcal{E}^\text{train}}

\newcommand{\eqs}{\mbox{$\epsilon_{q_s}$}}
\newcommand{\sqr}{\mbox{$s^0_{\qrho}$}}

\newcommand{\epsopt}{\epsilon^{\text{opt}}}

\newcommand{\bo}[1]{\mbox{$\mathbf{#1}$}}
\def\s{\bo{s}}
\def\x{\bo{x}}
\def\y{\bo{y}}
\def\X{\bo{X}}

\newcommand{\pmp}{\rho^\text{MP}}

\preprint{APS/123-QED}

\title{Statistical mechanics of high-dimensional inference}


\author{Madhu Advani}
\email{msadvani@stanford.edu}
 \author{Surya Ganguli}
 \email{sganguli@stanford.edu}
\affiliation{Department of Applied Physics, Stanford University, Stanford, CA 94305}


\begin{abstract}

To model modern large-scale datasets, we need efficient algorithms to infer a set of $P$ unknown model parameters from $N$ noisy measurements.  What are fundamental limits on the accuracy of parameter inference, given finite signal-to-noise ratios, limited measurements, prior information, and computational tractability requirements?  How can we combine prior information with measurements to achieve these limits? Classical statistics gives incisive answers to these questions as the measurement density $\alpha = \frac{N}{P}\rightarrow \infty$.   However, these classical results are not relevant to modern high-dimensional inference problems, which instead occur at finite $\alpha$.  We formulate and analyze high-dimensional inference as a problem in the statistical physics of quenched disorder.  Our analysis uncovers fundamental limits on the accuracy of inference in high dimensions, and reveals that widely cherished inference algorithms like maximum likelihood (ML) and maximum-a posteriori (MAP) inference cannot achieve these limits.  We further find optimal, computationally tractable algorithms that {\it can} achieve these limits.  Intriguingly, in high dimensions, these optimal algorithms become computationally simpler than MAP and ML, while still outperforming them.  For example, such optimal algorithms can lead to as much as a 20\% reduction in the amount of data to achieve the same performance relative to MAP.  Moreover, our analysis reveals simple relations between optimal high dimensional inference and low dimensional scalar Bayesian inference, insights into the nature of generalization and predictive power in high dimensions, information theoretic limits on compressed sensing, phase transitions in quadratic inference, and connections to central mathematical objects in convex optimization theory and random matrix theory.

\end{abstract}

\maketitle

\section{Introduction}

Remarkable advances in measurement technologies have thrust us squarely into the modern age of ``big-data,"  which yields the potential to revolutionize a variety of fields spanning the sciences, engineering, and humanities, including neuroscience \cite{Sejnowski2014,ganguli2012annrevs}, systems biology \cite{Clarke2008a}, health care \cite{Raghupathi2014}, economics \cite{Fan2011}, social science \cite{Leskovec2009}, and history \cite{Jockers2013}.  However, the advent of large scale data sets presents severe statistical challenges that must be solved if we are to gain conceptual insights from such data.

\par
\begin{figure}[b]
\begin{center}
\includegraphics[width=.48\textwidth]{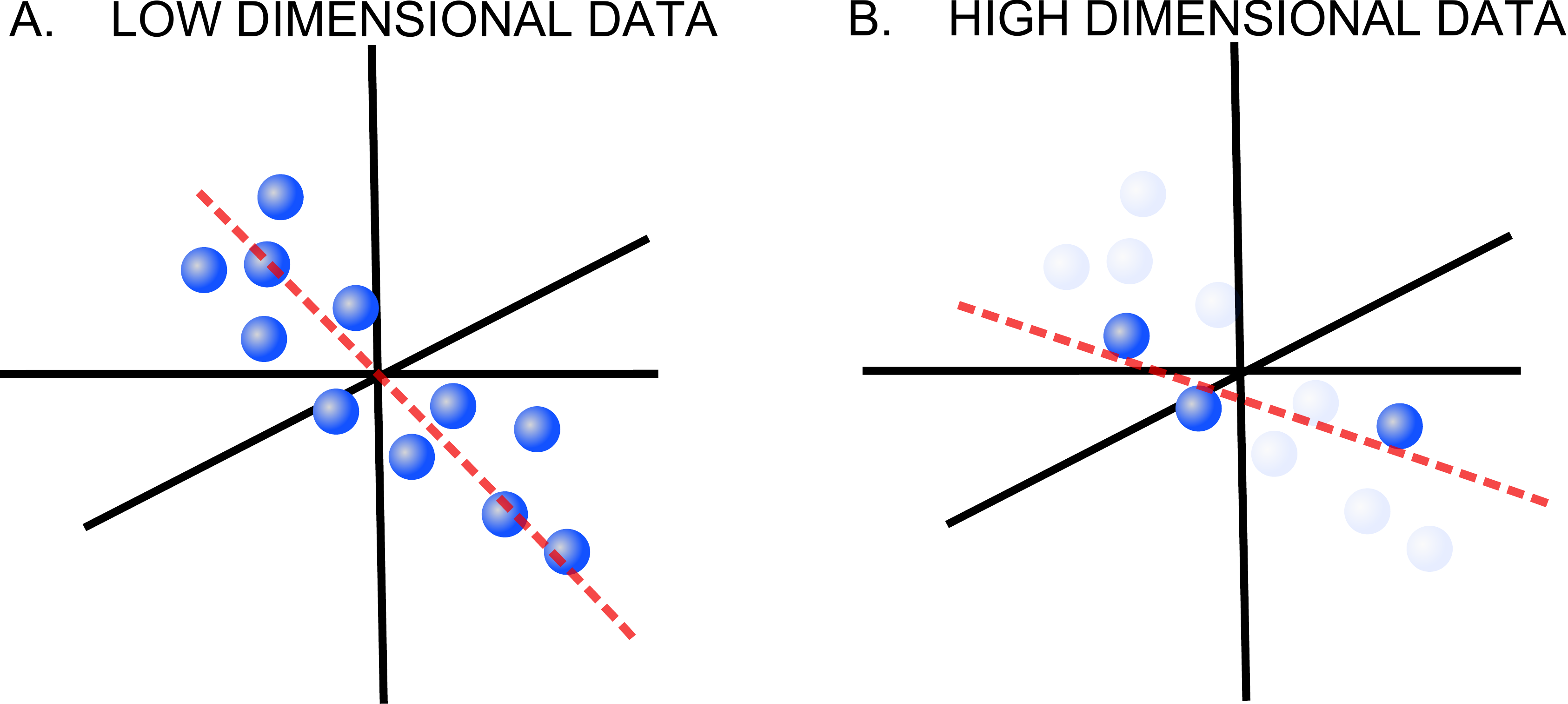}
\caption{A cartoon view of low (A) versus high (B) dimensional data.  In the latter scenario, a finite measurement density, or ratio between data points and dimensions, leads to errors in inference.
}
\label{fig:lowhigh}
\end{center}
\end{figure}

 A fundamental origin of the difficulty in analyzing many large scale data sets lies in their high dimensionality.   For example, in classically designed experiments, we often measure a small number of $P$ variables, chosen carefully ahead of time to test a specific hypothesis, and we take a large number of $N$ measurements.  Thus the measurement density $\alpha=\frac{N}{P}$ is extremely large, and such data sets are {\it low} dimensional: they consist of a large number of $N$ points in a low $P$ dimensional space (Fig. \ref{fig:lowhigh}A).  Much of the edifice of classical statistics operates within this low-dimensional, high measurement density limit.  Indeed, as reviewed below, as $\alpha \rightarrow \infty$, classical statistical theory gives us fundamental limits on the accuracy with which we can infer statistical models of such data, as well as the optimal statistical inference procedures to follow in order to achieve these limits.

\par In contrast to this classical scenario, our technological capacity for high-throughput measurements has led to a dramatic cultural shift in modern experimental design across many fields.  We now often simultaneously measure many variables at once in advance of choosing any specific hypothesis to test.  However, we may have limited time or resources to conduct such experiments, so we can only make a limited number of such simultaneous measurements.  For example, through multielectrode recordings, we can simultaneously measure the activity $P=1000$ neurons in mammalian circuits, but only for $N=O(100)$ trials of any given trial type.    Through microarrays, we can simultaneously measure the expression levels of $P=O(6000)$ genes in yeast, but again in a limited number of $N=O(100)$ experimental conditions.   Thus while both $N$ and $P$ are large, the measurement density $\alpha$ is finite. Such datasets are {\it high} dimensional, in that they consist of a small number of points in a high dimensional space (Fig. \ref{fig:lowhigh}B), and it can be extremely challenging to detect regularities in such data \cite{Advani2013}.  Moreover, classical statistical theory gives no prescriptions for how to optimally analyze such data.
\par In this work, we extend classical statistical theory to the modern age of high dimensional data, obtaining fundamental generalizations of statistical theorems dating back to the 1940's \cite{Cramer1946,Huber1973}.  We do so by interpreting the problem of high dimensional statistical inference within the framework of statistical physics.  In particular we focus on one of the most ubiquitous statistical inference procedures: regression, which attempts to find a linear relationship between a cloud of data points and another variable of interest.  By exploiting the methods of statistical mechanics, we obtain fundamental limits on the accuracy of high dimensional inference as well as the optimal procedures to follow to achieve these limits.  Our results reveal surprisingly simple connections between optimal high dimensional inference and low dimensional scalar Bayesian estimation, as well as quantitative insights into how the predictive power, or generalization capability, of an inference algorithm is related to its accuracy in separating signal from noise. Moreover, a variety of topics, including random matrix theory, compressed sensing, and fundamental objects in convex optimization theory, such as proximal mappings and Moreau envelopes, emerge naturally through our analysis.  We give an intuitive summary of our results in the discussion section.

\subsection{Statistical inference framework}

To more concretely introduce this work we give a precise definition of the problems we are solving.  Formally, let $\s^0$ be an unknown $P$ dimensional vector governing the linear response of a system's scalar output $y$ to a $P$ dimensional input $\x$ through the relation $y = \x \cdot \s^0 + {\epsilon}$, where $\epsilon$ denotes noise originating either from unobserved inputs or imperfect measurements.  For example, in sensory neuroscience, $y$ could reflect a linear approximation of the response of a single neuron to a sensory stimulus $\x$, so that $\s^0$ is the neuron's receptive field.  Alternatively, in genetic networks, $y$ could reflect the linear response of one gene to the expression levels $\x$ of a set of $P$ genes.  Suppose we perform $N$ measurements, indexed by $\mu=1,\dots,N$ in which we probe the system with an input $\x^\mu$ and record the resulting output $y^\mu$.  This yields a set of noisy measurements constraining the linear response vector $\s^0$ through the $N$ equations $y^\mu = \bo{x}^\mu \cdot \mathbf{s^0} + \epsilon^\mu$.

We assume the noise $\epsilon^\mu$ and components $\s^0_i$ are each drawn i.i.d. from a zero mean noise density $\peps(\epsilon)$, and a prior distribution $P_s(s)$. For convenience below, we define signal and noise energies in terms of the minus log probability of their respective distributions:  $E_\epsilon = - \log \peps$ and $E_s = -\log P_s$.  We further assume the experimental design of inputs is random: input components $\x^\mu_i$ are drawn i.i.d. from a zero mean Gaussian with variance $\frac{1}{P}$, yielding inputs of expected norm $1$.   In many systems identification applications, including for example in sensory neuroscience, this random design would correspond to a white-noise stimulus.  Now, given knowledge of the $N$ input-output pairs $\{ \x^\mu, y^\mu \}$, the noise density $\peps$, and the prior information encoded in $P_s$,  we would like to infer, in a computationally tractable manner, an estimate $\hat{\s}$ of the true response vector $\s^0$.  A critical parameter governing inference performance is the ratio of the number of measurements $N$ to the dimensionality $P$ of the unknown model parameter $\s^0$, i.e. the measurement density $\alpha = \frac{N}{P}$.

The performance of any inference procedure can be characterized in several ways.  Most simply, we would like to achieve a small, per-component
mean square error, $\qs = \frac{1}{P} \sum_{i=1}^P ( \hat{\s}_i - \s^0_i )^2$ in inferring the true parameters, or signal $\s^0$.   Alternatively, it is useful to note that any inference procedure yielding an estimate $\hat \s$ implicitly decomposes the measurement vector $\y$ into the sum of a signal component $\mathbf{X} \hat \s$ and a noise estimate $\hat{\boldsymbol{\epsilon}} = \mathbf{y} - \mathbf{X}{\hat \s}$.   Thus an inference procedure corresponds to a particular separation of measurements into estimated signal and noise,  $\mathbf{y} =  \mathbf{X}{\hat \s} + \hat{\boldsymbol{\epsilon}}$, which will generically differ from the true decomposition, $\mathbf{y} =  \mathbf{X}{\s^0} + \boldsymbol{\epsilon}$.  While $q_s$ reflects the error in estimating signal,  $q_\epsilon = \frac{1}{N} \sum_{\mu=1}^N ( \hat{\boldsymbol{\epsilon}}_\mu - {\boldsymbol{\epsilon}}_\mu)^2$ reflects the error in estimating noise.  Finally, one of the main performance measures of an inference procedure is its ability to generalize, or make predictions about the measurement outcome $y$ in response to a new randomly chosen input $\x$ not present in the training set $\{ \x^\mu \}$.   Given an estimate $\hat \s$, it can be used to make the prediction $\hat y =  \x \cdot \hat \s$, and the average performance of this prediction is captured by the generalization error $\Egen = \qav{(y - \hat y)^2}$.  Here the double average $\qav{\cdot}$ denotes an average over both the training data $\{\x^\mu, y^\mu \}$, which $\hat \s$ depends on, and the held out testing data $\{\x, y \}$, which is necessarily independent of $\hat \s$.  An alternate measure of performance is the average error in the ability of $\hat \s$ to simply predict the training data: $\Etrain = \frac{1}{N}\sum_{\mu=1}^N{(y^\mu - \x^\mu\cdot\hat{\s})^2} = \frac{1}{N} \sum_{\mu=1}^N{\hat{\boldsymbol{\epsilon}}_\mu^2}$.  In general, $\Etrain < \Egen$,  since through the process of inference,  the learned parameters $\mathbf{\hat s}$ can acquire subtle correlations with the particular realization of training inputs $\{ \mathbf{\x^\mu} \}$ and noise $\{ \epsilon^\mu \}$ so as to reduce $\Etrain$.  Situations where $\Etrain \ll \Egen$ correspond to inference procedures that overfit to the training data, and do not exhibit predictive power by generalizing to new data.

Now what inference procedures can achieve good performance in a computationally tractable manner?  Regularized M-estimation (see \cite{van2000asymptotic,Huber2009} for reviews) yields a large family of computationally tractable estimation procedures in which $\hat{\s}$ is computed through the minimization
\begin{equation}
\label{mEstEq}
\mathbf{\hat{\s}} = \argmin_{\mathbf{s}} {\left[\sum_{\mu=1}^N{\rho(y^\mu - \mathbf{x}^{\mu} \cdot \mathbf{s})} + \sum_{i=1}^P{\sigma(s_i)}\right]}.
\end{equation}
Here $\s$ is a candidate response vector, $\rho$ is a loss function that penalizes deviations between actual measurements $y^\mu$ and expected measurements $\x^\mu \cdot \s$ under the candidate $\s$, and $\sigma(s)$ is a regularization function that exploits prior information about $\s^0$.

In the absence of such prior information, a widely used procedure is maximum likelihood (ML) inference,
\begin{equation}
\sml = \argmax_s \, \log P \left( \{ y^\mu \} \, | \,  \{ \x^\mu \} , \s  \right).
\end{equation}
ML corresponds to noise energy minimization through the choice $\rho = E_\epsilon$ and $\sigma=0$  in \eqref{mEstEq}.  Amongst all unbiased estimation procedures (in which $\langle \hat{\s} \rangle = \s^0$, where $\langle \cdot \rangle$ denotes an average over noise realizations), this energy minimization is optimal, but only in the low dimensional limit. Thus, amongst unbiased procedures,  ML achieves the minimum mean squared error (MMSE),  when $\alpha \rightarrow \infty$, but not at finite $\alpha$.  Recent work \cite{Donoho2013,Karoui2013,Bean2013} uses non-statistical mechanics based methods to find the optimal $\rho$ at finite $\alpha$, but leaves open the fundamental question of how to optimally exploit prior information by choosing  a nonzero $\sigma$.
\par With prior knowledge, the Bayesian posterior mean achieves the MMSE estimate,
\begin{equation}
\smmse = \avg{ \s \, | \, \{ y^\mu, \x^\mu \}}  = \, \int \text{d}\s \, \s \, P \left( \s \, | \, \{ y^\mu, \x^\mu \} \right).
\end{equation}
However, while no inference procedure can outperform high dimensional Bayesian inference of the posterior mean, this procedure is not an M-estimator, and it is often computationally intractable due to the $P$ dimensional integral.  A widely used, more computationally tractable surrogate is maximum {\it a-posteriori} (MAP) inference,
\begin{equation}
\smap = \argmax_s \, \log P \left( \s \, | \, \{ y^\mu, \x^\mu \} \right),
\end{equation}
which corresponds to noise and signal energy minimization through the choice $\rho = E_\epsilon$ and $\sigma = E_s$ in \eqref{mEstEq}. MAP inference, by potentially introducing a non-zero bias (so that  $\langle \hat{\s} \rangle \neq \s^0$) can out-perform ML at finite $\alpha$, but is not in general optimal.  However, the exploitation of prior information through a judicious, even if suboptimal, choice of $\sigma$ can dramatically reduce estimation error.  For example, the seminal advance of compressed sensing (CS)  \cite{donoho2003optsparse,emmanuel2006candes,candes2005decoding} uses $\rho=\frac{1}{2}\epsilon^2$ and $\sigma \propto |s|$. This choice can lead to accurate inference of sparse $\s^0$ even when $\alpha < 1$,  where sparsity means that $P_s(s)$ assigns a small probability to nonzero values.

\par Despite the important and successful special cases of MAP inference and CS, there exists no general method to choose the best $\rho$ and $\sigma$ for inference.
The central questions we address in this work are: (1) Given an estimation problem defined by the triplet of measurement density, noise and prior ($\alpha$, $E_\epsilon$, $E_s$), and an estimation procedure defined by the loss and regularization pair  ($\rho$, $\sigma$), what is the typical error $q_s$ achieved for random inputs $\x^\mu$ and noise $\epsilon^\mu$? (2)  What is the minimal achievable estimation error $\qopt$ over all possible choices of convex procedures ($\rho$, $\sigma$)? (3) Which procedure ($\rhoopt$, $\sigopt$) achieves the minimal error $\qopt$? (4) Are there simple universal relations between $q_s$ and $q_\epsilon$ which measure the ability of an inference procedure to accurately separate signal and noise, and $\Etrain$ and $\Egen$, which capture the predictive power of an inference procedure? Our discussion section gives a summary of the answers we find to these questions.
%


%

\section{Results}

\subsection{Review and formulation of classical scalar inference}

Before considering the finite $\alpha$ regime, it is useful to review classical statistics in the $\alpha \rightarrow \infty$ limit, in the context of scalar estimation, where  $P=1$.  In particular, we formulate these results in a suggestive manner that will aid in understanding the novel phenomena that emerge in modern, high dimensional statistical inference, derived below.  Here, for simplicity, we choose the scalar measurements $x^\mu = 1 \, \forall \, \mu$ in \eqref{mEstEq}.  Thus we must estimate the scalar $s^0$ from $\alpha = N$ noisy measurements, $y^\mu = s^0 + \epsilon^\mu$.  With no regularization ($\sigma=0$), for large $N$, $\hat s$ in \eqref{mEstEq} will be close to $s^0$, so simply Taylor expanding $\rho$ about $s^0$ yields the asymptotic error (see, \cite{van2000asymptotic,Huber2009}, \cite{PRLsupp}, appendix A.1)
\begin{equation}
\qs = \frac{1}{N} \frac{\qav{\rho'(\epsilon)^2}_{\epsilon}}{\qav{\rho''(\epsilon)}^2_{\epsilon}}.
\label{singleParVar}
\end{equation}

The Cramer-Rao (CR) bound is a fundamental information theoretic lower bound, at any $N$, on the error of {\it any} unbiased estimator $\hat s( \{ y^\mu \} )$ (obeying  $\langle \hat s - s^0 \rangle_\epsilon = 0$):
\begin{equation}
\qs\ge \frac{1}{N}\frac{1}{\fish{\epsilon}},
\label{eq:crbound}
\end{equation}
where $\fish{\epsilon}$ is the Fisher information from a single measurement $y$,
\begin{equation}
\fish{\epsilon} = \qav{\left(\frac{\partial}{\partial s^0}\log P(y \, | \, s^0)\right)^2}_{y} = \qav{ \left( \frac{\partial }{\partial \epsilon} E_\epsilon \right)^2}_{\epsilon}.
\end{equation}
The Fisher information measures the susceptibility of the output $y$ to small changes in the parameter $s^0$.  The higher this susceptibility, the lower the achievable error in \eqref{eq:crbound}.  For finite $N$, it is not clear there exists a loss function $\rho$ whose performance saturates the CR bound.  However, a central result in classical statistics states that as $N \rightarrow \infty$, the choice $\rho = E_\epsilon$ saturates \eqref{eq:crbound}, as can be seen by substituting $\rho = E_\epsilon$ in \eqref{singleParVar} (\cite{PRLsupp}, appendix A.2).

\par  With knowledge of the true signal distribution $P(s^0)$, the posterior mean  $\avg{ s \, | \, \{ y^\mu \}}  = \int \text{d}s \, s \, P \left( \s \, | \, \{ y^\mu \} \right)$ achieves minimal possible error $q_s$, amongst all inference procedures, biased or not, at any finite $N$.  We compute this minimal $q_s$, in the limit of large $N$, via a saddle point approximation to this Bayesian integral, yielding a mean field theory (MFT) for low dimensional Bayesian inference  (\cite{PRLsupp}, appendix A.3), where the $N$ measurements $y^\mu$ of $s^0$, corrupted by {\it non}-Gaussian noise $\epsilon^\mu$, can be replaced by a {\it single} measurement $y = s^0 +\sqrt{q_d}z$, corrupted by an effective Gaussian noise of variance
\begin{equation}
q_d = \frac{1}{N J[\epsilon]}.
\label{eq:qdnoiseeff}
\end{equation}
Here $z$ is a zero mean unit variance Gaussian variable.  In our MFT, $q_s$ is the MMSE error $q_s^\text{MMSE}$ of this equivalent single
measurement, Gaussian noise inference problem:
\begin{equation}
q_s^\text{MMSE}(q_d) = \qav{ \left(s^0 - \avg{s \, | \, y = s^0 + \sqrt{q_d} z} \right)^2 }_{s^0,z}.
\label{eq:qmmse}
\end{equation}
We further prove a general lower bound on the asymptotic error
\begin{equation}
q_s  \geq \frac{1}{N\fish{\epsilon} + \fish{s^0}},
\label{eq:bayesbound}
\end{equation}
and demonstrate that this bound is tight when the signal and noise are Gaussian (\cite{PRLsupp}, appendix A.3).
\par Thus, the classical theory of unbiased statistical inference as the measurement density $\alpha \rightarrow \infty$ reveals that ML achieves information theoretic limits on error \eqref{eq:crbound}.  Moreover, our novel asymptotic analysis of Bayesian inference as $\alpha \rightarrow \infty$ (Eqs. \ref{eq:qdnoiseeff}-\ref{eq:bayesbound}), reveals the extent to which biased procedures that optimally exploit prior information can circumvent such limits.  Our work below constitutes a fundamental extension of these results to modern high dimensional problems at finite measurement density.

\begin{figure}[t]
\begin{center}
\includegraphics[width=.48\textwidth]{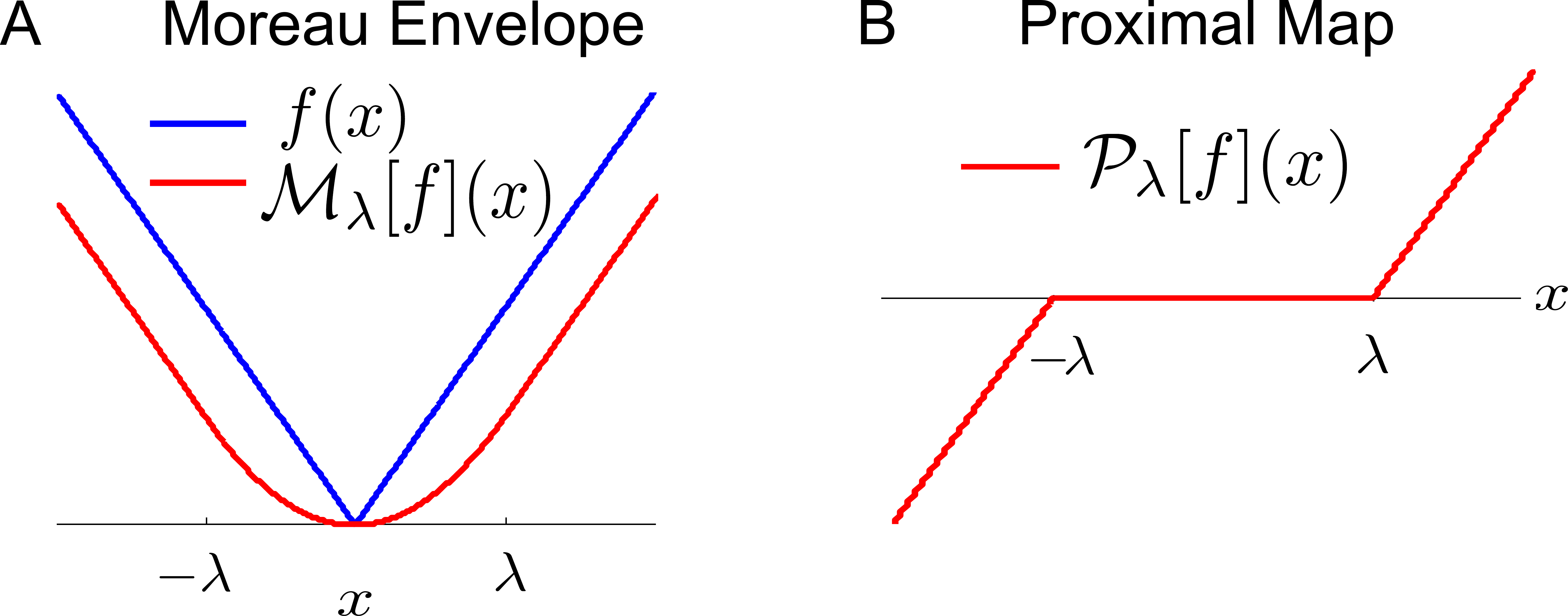}
\caption{(A) An example of a smooth, lower bounding Moreau envelope $\mor{\lambda}{f}{x}$ in \eqref{eq:moreau} for $f(x) = |x|$.
Explicitly, $\mor{\lambda}{f}{x} = \frac{x^2}{2\lambda}$ for $|x| \leq \lambda$, and $|x| - \frac{\lambda}{2}$ for $|x| \geq \lambda$.
(B) The proximal map $\proxop{\lambda}{f}{x}$ in \eqref{eq:proxmap} for $f(x) = |x|$.  Explicitly, $\proxop{\lambda}{f}{x}=0$ for
$|x| \leq \lambda$, and $x - \text{sign}(x) \lambda$  for $|x| \geq \lambda$.
Thus the proximal descent map $x \rightarrow \proxop{\lambda}{f}{x}$ moves $x$ towards the minimum of $f(x)$.
 }
\label{fig:prox_mor}
\end{center}
\end{figure}



\subsection{Statistical mechanics framework}
To understand the properties of the solution $\hat \s$ to \eqref{mEstEq}, we define an energy function
\begin{equation}
E(\s) = \sum_{\mu=1}^N{\rho\left(y^\mu - \mathbf{x^\mu} \cdot \s \right)} + \sum_{i=1}^P{\sigma(s_i)},
\label{energy}
\end{equation}
yielding a Gibbs distribution $P_G(\s) = \frac{1}{Z} e^{-\beta E(\s)}$ that freezes onto the solution of \eqref{mEstEq} in the zero temperature $\beta \rightarrow \infty$ limit.  In this statistical mechanics system, $\x^\mu$, $\epsilon^\mu$ and $\s^0$ play the role of quenched disorder, while the components of the candidate parameters $\s$ comprise thermal degrees of freedom.  For large $N$ and $P$, we expect self-averaging to occur: the properties of $P_G$ for any typical realization of disorder coincide with the properties of $P_G$ averaged over the disorder.   Therefore we compute the average free energy $-\beta \bar F \equiv \qav {\text{ln}\, Z}_{\bo{x}^\mu, \epsilon^\mu, \bo{s}^0}$ using the replica method \cite{Mezard1987}.  We employ the replica symmetric (RS) approximation, which is effective for convex $\rho$ and $\sigma$.  Interestingly, our calculation (\cite{PRLsupp}, section 2.1) goes through without assuming a quadratic loss, as in previous replica analyses of compressed sensing \cite{Rangan2009,Ganguli2010}. For a review of statistical mechanics methods applied to high dimensional inference in diverse settings, see \cite{Advani2013}.




\begin{figure*}[t!]
\begin{center}
\includegraphics[width=0.7\textwidth]{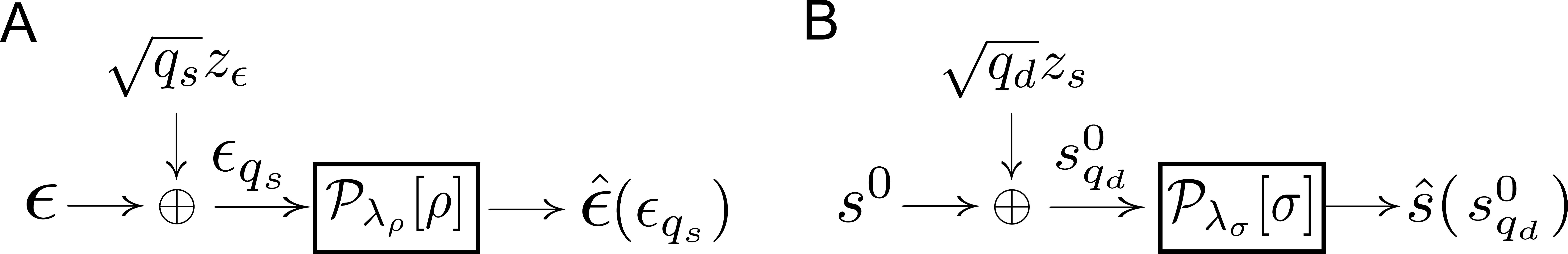}
\caption{A low-dimensional scalar MFT for high dimensional inference. (A) and (B) are schematic descriptions of Eqns.  \eqref{eq:addnoise} and \eqref{eq:rsprox}.  They describe a pair of  scalar statistical estimation problems, one for a noise variable $\epsilon$, drawn from $P_\epsilon$ in (A), and the other for a signal variable $s^0$, drawn from $P_s$ in (B).   Each variable is corrupted by additive Gaussian noise, and from these noise corrupted measurements, the original variables are estimated through proximal descent steps, yielding a noise estimate $\hat \epsilon$ in (A) and a signal estimate $\hat s$ in (B).  The MFT distributions $\pmf(\epsilon, \hat \epsilon)$ and $\pmf(s^0, \hat s)$ are obtained by integrating out $\ze$ and $\zs$ in (A) and (B) respectively.  These joint MF distributions describe the joint distribution of pairs of single components $(\epsilon_\mu, \hat \epsilon_\mu)$, and $(s^0_i, \hat s_i)$  in \eqref{mEstEq}, after integrating out all other elements of the quenched disorder in the training data and true signal.}
\label{fig:schema}
\end{center}
\end{figure*}
\par Central objects in optimization theory emerge naturally from our replica analysis, and the resulting mean field theory (MFT) is most naturally described in terms of them.  First is the proximal map $x \rightarrow \proxop{\lambda}{f}{x}$, where
\begin{equation}
\proxop{\lambda}{f}{x} = \argmin_y\left( \frac{(y-x)^2}{2\lambda} + f(y) \right).
\label{eq:proxmap}
\end{equation}
This mapping is a {\it proximal descent step} that maps $x$ to a new point that minimizes $f$, while remaining proximal to $x$, as determined by a scale $\lambda$.  The proximal map is closely related to the Moreau envelope of $f$,  given by
\begin{equation}
\mor{\lambda}{f}{x} = \min_y\left( \frac{(y-x)^2}{2\lambda} + f(y) \right).
\label{eq:moreau}
\end{equation}
\mornoarg{\lambda}{f} is a minimum convolution of $f(x)$ with a quadratic $\frac{x^2}{2\lambda}$, yielding a lower bound on $f$ that is smoothed over a scale $\lambda$.  See Fig. \ref{fig:prox_mor}AB for an example.  The proximal map and Moreau envelope are related:
\begin{equation}
\proxop{\lambda}{f}{x} = x - \lambda \, \morp{\lambda}{f}{x},
\label{eq:proxmorrel}
\end{equation}
where the prime denotes differentiation w.r.t. $x$. Thus a proximal descent step on $f$ can be viewed as a gradient descent step on $\mornoarg{\lambda}{f}$ with step length $\lambda$.  See \cite{PRLsupp}, appendix C.1, and also \cite{Parikh2013} for a review of these topics.
\par Our replica analysis yields a pair of zero temperature MFT distributions $\pmf(s^0, \hat s)$ and $\pmf(\epsilon, \hat \epsilon)$. The first describes the joint distribution of a single component $(s^0_i, \hat s_i)$ in \eqref{mEstEq}, while the second describes the joint distribution of a noise component $\epsilon^\mu$ and its estimate $\hat \epsilon^\mu \equiv y^\mu - \x^\mu \cdot \mathbf{\hat \s}$.  The MFT distributions can be described in terms of a pair of coupled scalar noise and signal estimation problems, depending on a set of RS order parameters ($\qs$, $\qrho$, $\lrho$, $\lsig$).  Here $\qs$ and $\qrho$ reflect the variance of additive Gaussian noise that corrupts the noise $\epsilon$ and signal $s^0$, respectively, yielding the measured variables,
\begin{equation}
\eqs = \epsilon + \sqrt{\qs} \, \ze \qquad   \sqr = s^0 + \sqrt{\qrho} \, \zs,
\label{eq:addnoise}
\end{equation}
where $\ze$ and $\zs$ are independent zero mean unit variance Gaussians.  From these measurements, estimates $\hat \epsilon$ and $\hat s$ of the original noise $\epsilon$ and signal $s^0$ are obtained through proximal descent steps on the loss $\rho$ and regularization $\sigma$:
\begin{equation}
\hat \epsilon(\eqs) = \proxop{\lambda_\rho}{\rho}{\epsilon_{\qs}} \qquad  \hat s(\sqr) = \proxop{\lambda_\sigma}{\sigma}{s^0_{\qrho}},
\label{eq:rsprox}
\end{equation}
where $\lrho\,$  and $\lsig\,$ reflect scale parameters.
The joint MFT distributions are then obtained by integrating out $z_\epsilon$ and $z_s$.  These MFT equations can be thought of as defining a pair scalar estimation problems, one for the noise, and one for the signal (see Fig. \ref{fig:schema}AB for a schematic).
\par The order parameters obey self-consistency conditions that couple the performance of these scalar estimation problems:
\begin{equation}
\qrho = \frac{\qav{\morp{\lambda_\rho}{\rho}{\eqs}^2}_{\eqs}}{\alpha\qav{\morpp{\lambda_\rho}{\rho}{\eqs}}^2_{\eqs}} \,
\quad \qs = \qav{ (\hat s - s^0 )^2 }_{\sqr},
\label{rsreg1}
\end{equation}
\begin{equation}
1 - \frac{1}{\alpha} \frac{\lrho}{\lsig} = \qav{\hat \epsilon'(\eqs) }_{\eqs} \,
\quad \frac{\lrho}{\lsig} = \qav{ \hat s'(\sqr) }_{\sqr}.
\label{rsreg2}
\end{equation}
%
Here \qav{\cdot} denotes averages over the quenched disorder in \eqref{eq:addnoise}.  The pair of MF distributions determine various measures of inference performance in \eqref{mEstEq}.  In particular, $\qs\,$  predicts the typical per-component error of the learned model parameters, or signal $\hat\s$, while $q_\epsilon = \qav{ (\hat \epsilon - \epsilon )^2 }_{\eqs}$ predicts the typical per-component error of the estimated noise.  The model's prediction, or generalization error $\Egen = \qav{(y - \x\cdot\hat{\s})^2}$ on a new example  $(\x,y)$ not present in the training set $\{\x^\mu, y^\mu\}$ can be obtained by substituting $y = \x\cdot\s^0 + \epsilon$ into $\Egen$. This yields the MFT prediction for the generalization error, $\Egen = \qav{(\eqs)^2} = q_s + \avg{\epsilon^2}$.  In contrast, the MFT prediction for the training error is simply $\Etrain = \qav{\hat \epsilon(\eqs)^2}$.

\par Because the proximal map is contractive, with Jacobian less than 1 \cite{Parikh2013},  the MFT predicts, as expected, that $\Etrain < \Egen$.  The reason for the reduced $\Etrain$ is due to the subtle correlations that the learned parameters $\mathbf{\hat s}$ can acquire with the particular realization of training inputs $\{ \mathbf{\x^\mu} \}$ and noise $\{ \epsilon^\mu \}$, through the optimization in \eqref{mEstEq}. Remarkably, these subtle correlations are captured in the MFT simply through a proximal descent step in \eqref{eq:rsprox} on the cost $\rho$.  This step contracts the variable $\eqs$ controlling $\Egen$ towards the minimum of $\rho$ at the origin, leading to smaller $\Etrain$.  We explore many more consequences of this MFT below.

\subsection{Inference without prior information}
\begin{figure*}[t!]
\begin{center}
\includegraphics[width=0.75\textwidth]{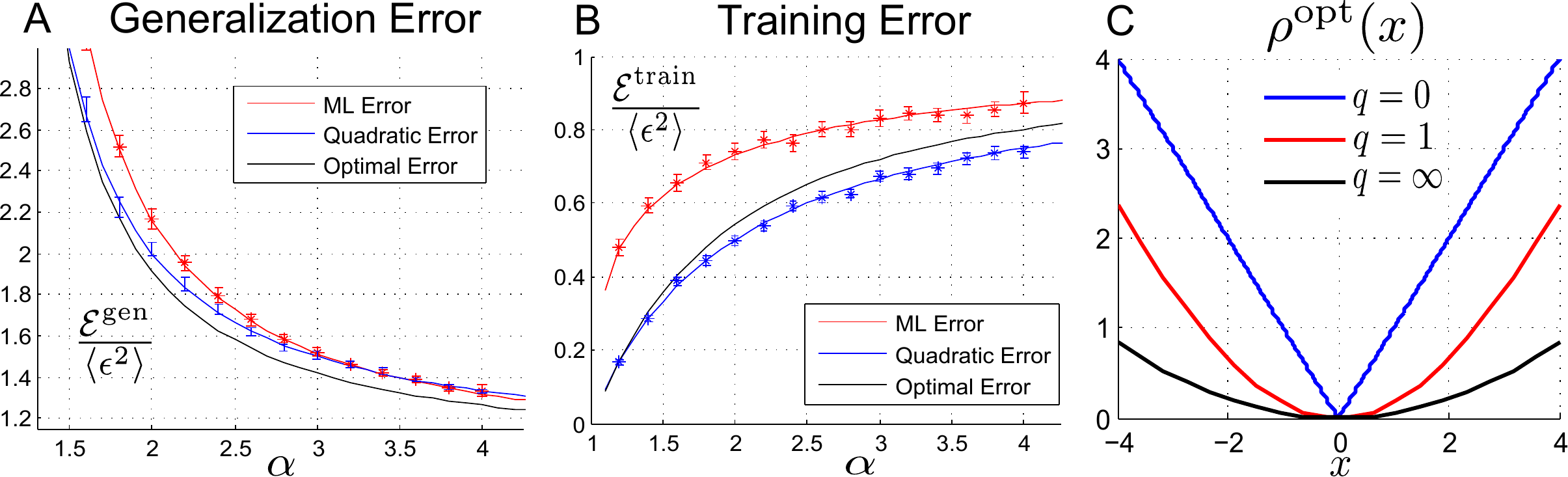}
\caption{Unregularized inference for Laplacian noise $E_{\epsilon}  = |\epsilon|$.  A comparison of the generalization error (A) and training error (B) of the optimal unregularized M-estimator \eqref{eq:optRhoUnreg} (black) with ML (red), and quadratic (blue) loss functions.  Solid curves reflect theoretically derived predictions of performance.  Error bars reflect performance obtained through numerical optimization of \eqref{mEstEq} using standard convex optimization solvers for finite size problems ($N$ and $P$ vary, with $N = \alpha P$ and $\sqrt{NP} = 250$).   The width of the error bars reflect standard deviation of performance across $100$ different realizations of the quenched disorder.  (C) The shape of the optimal loss function in (\ref{eq:optRhoUnreg}) for high dimensional inference as a function of the error or smoothing parameter $q$. As $\alpha$ varies from high to low measurement density, $q$ varies from low to high values, and the optimal loss function varies from the ML loss to quadratic.  Intermediate versions of the optimal loss behave like a smoothed version of the ML loss, with increased smoothing as measurement density decreases (or dimensionality increases).}
\label{q0plot}
\end{center}
\end{figure*}
If we cannot exploit prior information, we simply choose $\sigma=0$, which yields $\hat s = s^0_{\qrho}$ in \eqref{eq:rsprox}, so that the RHS of \eqref{rsreg1} and \eqref{rsreg2} reduce to  $\qs = \qrho$ and $\lrho = \lsig$.  Then, replacing $q_d$ with $q_s$ on the LHS of \eqref{rsreg1}, and comparing to \eqref{singleParVar}, we see that the high dimensional inference error is analogous to the low dimensional one with the number of measurements $N$ replaced by the measurement density $\alpha$,  the cost $\rho(\cdot)$ replaced by its Moreau envelope $\mor{\lambda_\rho}{\rho}{\cdot}$, and the noise $\epsilon$ further corrupted by additive Gaussian noise of variance $q_s$, with $q_s$ and $\lambda_\rho$ determined self-consistently through \eqref{rsreg1}-\eqref{rsreg2}.
\par As a simple example, consider the ubiquitous case of quadratic cost: $\rho(x) = \frac{1}{2}x^2$.  Then the proximal map \eqref{eq:rsprox} is simply linear shrinkage to the origin, $\hat \epsilon(\eqs) = \frac{1}{1+\lambda_\rho} \epsilon_{\qs}$, and \eqref{rsreg1} and \eqref{rsreg2} are readily solved: $\qs = \frac{1}{\alpha-1} \langle \epsilon^2 \rangle $,  $\lambda_\rho = \frac{1}{\alpha-1}$, yielding $\Egen =  \frac{\alpha}{\alpha-1}\langle \epsilon^2 \rangle$ and $\Etrain = \frac{\alpha-1}{\alpha}\langle \epsilon^2 \rangle$.  Thus as the measurement density approaches $1$ from above, the error in inferred parameters $\hat \s$ and $\Egen$ diverge, while $\Etrain$ vanishes, indicating severe overfitting.

\par Now, in the space of all convex costs $\rho$, for a given density $\alpha$ and noise energy $E_\epsilon$, what is the minimum possible estimation error $\qopt$?  By performing a functional minimization of $\qs$ over $\rho$  subject to the constraints  \eqref{rsreg1} and \eqref{rsreg2} ((see \cite{PRLsupp} sec. 4.1 and 5.1 for details) we find that $\qopt$ is the minimal solution to
\begin{equation}
\quad \qopt = \frac{1}{\alpha}\frac{1}{\fish{\epsilon_{\qopt}}} \ge \frac{1}{(\alpha-1)\fish{\epsilon}},
\label{optConNB}
\end{equation}
where the second inequality follows from the convolutional Fisher inequality (\cite{PRLsupp}, appendix B.2).  This result is the high dimensional analog of the Cramer-Rao bound in \eqref{eq:crbound}.  By the data processing inequality for Fisher information, $\fish{\epsilon_{\qopt}} < \fish{\epsilon}$, indicating higher error in the high dimensional \eqref{optConNB} than low dimensional setting \eqref{eq:crbound}. Thus the price paid for even optimal high-dimensional inference at finite measurement density, relative to ML inference at infinite density, is increased error due to the presence of additional gaussian noise with dimensionality dependent variance $q_s$.
\par Now can this minimal error $q^{opt}$ be achieved, and if so, which cost function $\rhoopt$ achieves it?  Constrained functional optimization over $\rho$ yields the functional equation $\mor{\qopt}{\rho}{x} = E_{\epsilon_{\qopt}}$ (see \cite{PRLsupp} sec. 5.1 for details), which can be inverted (see \cite{PRLsupp} appendix B.2) to find
\begin{equation}
\rhoopt(x)=-\mor{\qopt}{-E_{\epsilon_{\qopt}}}{x}.
\label{eq:optRhoUnreg}
\end{equation}
The validity of this equation under the RS assumption requires that $\rhoopt$ be convex.  Convexity of the noise energy $E_\epsilon$ is sufficient to guarantee the convexity of $\rhoopt$, and so for this class of noise, \eqref{eq:optRhoUnreg} yields the optimal inference procedure.
\par In the classical $\alpha \ra \infty$ limit, we expect $\qopt$ to be small; indeed to leading order in $\frac{1}{\alpha}$,  \eqref{optConNB}  has the solution $\qopt = \frac{1}{\alpha}\frac{1}{J[\epsilon]}$, while \eqref{eq:optRhoUnreg} reduces to $\rhoopt = E_\epsilon$, recovering the optimality of ML and its performance \eqref{eq:crbound} at infinite measurement density.
In the high dimensional $\alpha \rightarrow 1$ limit, $\qopt$ diverges, so that $\epsilon_{\qopt}$ approaches a Gaussian with variance $\langle \epsilon^2 \rangle + \qopt$, yielding in \eqref{eq:optRhoUnreg} $\rhoopt(x) = \frac{x^2}{2}$.  Thus, remarkably, at low measurement density, simple quadratic minimization, independent of the noise distribution, becomes an optimal inference procedure.  As the measurement density decreases, $\rhoopt$ interpolates between $E_\epsilon$ and a quadratic; in essence $\rhoopt$ at finite density $\alpha$ is a smoothed version of the ML choice $\rho = E_\epsilon$ where the amount of smoothing increases, as the density decreases (or dimensionality increases).  See Fig. \ref{q0plot} for an example of a family of optimal inference procedures, and their performance advantage relative to ML, for Laplacian noise ($E_\epsilon = | \epsilon |$).
\par These results are consistent with and provide a new statistical mechanics based derivation of results in \cite{Donoho2013,Karoui2013,Bean2013}, and they illustrate the severity of overfitting in the face of limited data.

\subsection{Inference with prior information}
We next explore how we can combat overfitting by optimally exploiting prior information about the distribution of the model parameters, or signal $s^0$.

\subsubsection{Optimal quadratic inference: a high SNR phase transition}

To understand the MFT for regularized inference, it is useful to start with the oft-used quadratic loss and regularization: $\rho(x) = \frac{1}{2}x^2$ and $\sigma(x) = \frac{1}{2} \gamma x^2$.  In this case, the proximal maps in \eqref{eq:rsprox} become linear and the RS equations \eqref{rsreg1} and \eqref{rsreg2} are readily solved (\cite{PRLsupp}, sec. 3.1).  It is useful to express the results in terms of the fraction of unexplained variance $\bar q_s  = \frac{q_s}{\langle s^2 \rangle}$ and the $\text{SNR}  = {\langle s^2 \rangle}/{\langle \epsilon^2 \rangle}$.  For quadratic inference, $\bar q_s$ depends on the signal and noise distributions only through the SNR.  We find that in the strong regularization limit, $\gamma \rightarrow \infty$, $\bar q_s \rightarrow 1$, as the regularization pins the estimate $\hat \s$ to the origin, while in the weak regularization limit $\gamma \rightarrow 0$, $\bar q_s \rightarrow \frac{1}{\text{SNR}(\alpha-1)}$, recovering the unregularized case.  There is an optimal intermediate value of the regularization weight, $\gamma = \frac{1}{\text{SNR}}$, leading to the highest fraction of variance explained.  Thus optimal quadratic inference obeys the principle that high-quality data, as measured by high SNR, requires weaker regularization.  For this optimal $\gamma$, $\bar q_s$ arises as the solution to the
\begin{figure*}[t!]
\begin{center}
\includegraphics[width=0.6\textwidth]{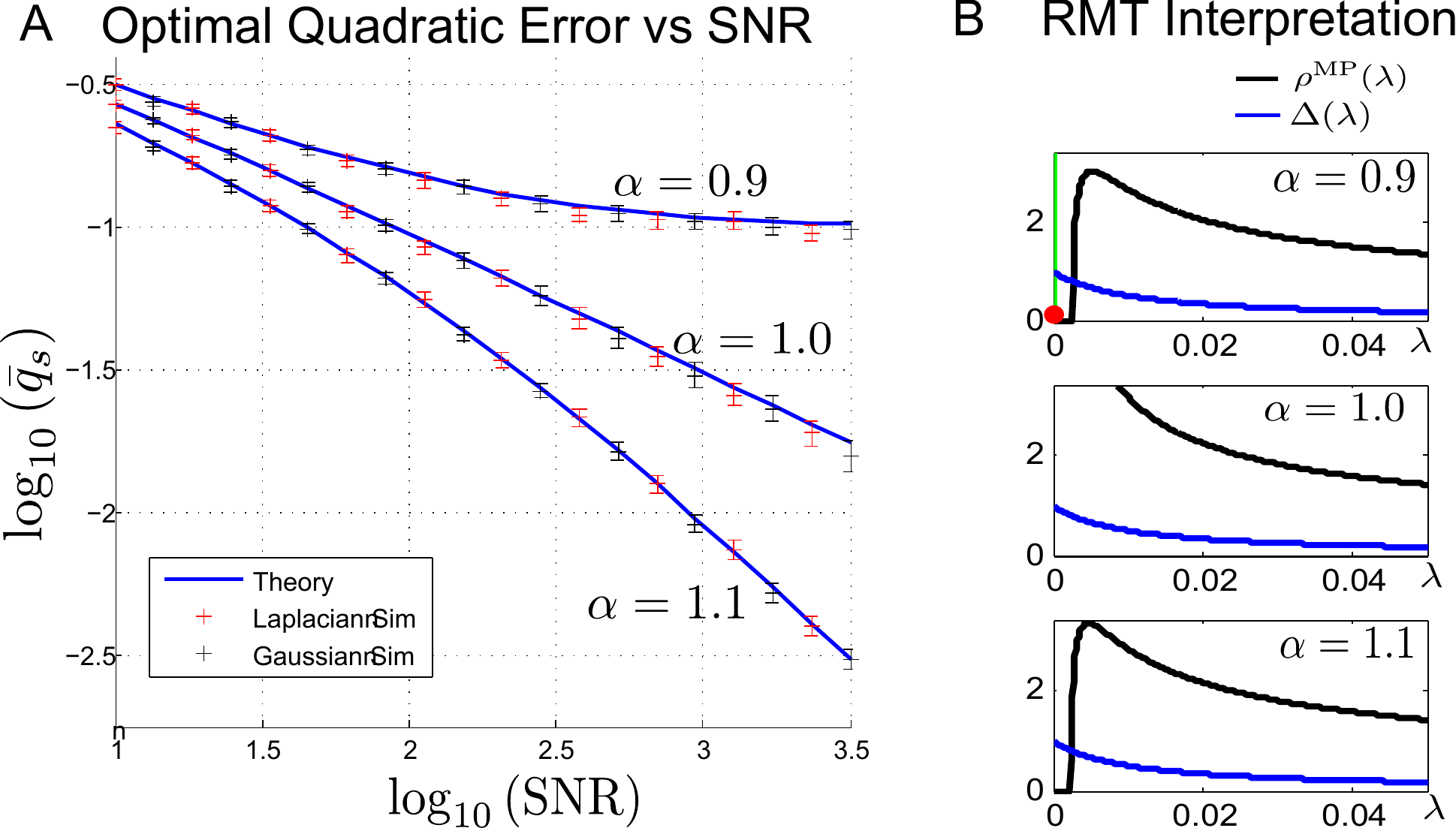}
\caption{A high SNR phase transition in optimal quadratic inference.  (A) At large SNR, the MSE of optimal quadratic inference exhibits three distinct scaling regimes for $\alpha<1, \alpha=1$, and $\alpha >1$ (see eq. \eqref{snrinf_case}), independent of the signal and noise distributions.   For example, when $\alpha =0.9 < 1$, $\bar q^{\text{Quad}}_s$ approaches a constant, whereas when $\alpha = 1$ or $\alpha = 1.1>1$, $\bar q^{\text{Quad}}_s$ approaches $0$ as $\snr^{-1/2}$ or $\snr^{-1}$ respectively. The theoretical curves (blue) match numerical experiments (error bars) for a finite sized problems ($N$ and $P$ vary with $N=\alpha P$ and $\sqrt{N P} = 300)$, where the error bars reflect the standard deviation across $80$ trials using both signal and noise either Gaussian (black)  or  Laplacian (red) distributed. (B) The behavior of the MP density (black) in \eqref{eq:marpas}.  For $\alpha \neq 1$ the nonzero continuous part of the density exhibits a gap at the origin, whereas for $\alpha=1$ the gap vanishes and the distribution diverges at the origin. For $\alpha < 1$ there is an additional $\delta$-function at the origin (green bar) with weight $1-\alpha$ (red dot).  The blue curve shows the function $\Delta(\lambda) = (1 +\lambda \cdot \snr )^{-1}$ appearing in the integral for $\bar q^{\text{Quad}}_s$ in \eqref{eq:qsbintegral}, for the value $\text{SNR} = 100$.}
\label{quadplot}
\end{center}
\end{figure*}
set of simultaneous equations
\begin{equation}
q_d = \frac{\langle \epsilon^2 \rangle + q_s}{\alpha} \qquad \qquad \frac{q_s}{\langle s^2 \rangle} =  \frac{1}{1 + \frac{\langle s^2 \rangle}{q_d}}.
\label{eq:qdqsquad}
\end{equation}
We denote the solution to these equations by $\bar q_s = \bar q^{\text{Quad}}_s( \alpha, SNR)$. This function is simply the fraction of unexplained variance of optimal quadratic inference at a given measurement density and SNR,  and an explicit expression is given by
\begin{equation}
\bar q^{\text{Quad}}_s = \frac{1 - \alpha - \phi + \sqrt{(\phi + \alpha -1)^2 + 4 \phi }}{2},
\label{eq:qsquadopt}
\end{equation}
where $\phi=\frac{1}{\text{SNR}}$ (see \cite{PRLsupp}, sec. 3.2 for details).
\par This expression simplifies in several limits.  At high $\snr \gg 1$,
\begin{align}
 \bar{q}^\text{Quad}_s =
\begin{cases}
1-\alpha &\alpha < 1,\\
\frac{1}{\sqrt{\snr}} & \alpha =1\\
\frac{1}{\snr (\alpha -1)}  &\alpha > 1.
\end{cases}
\label{snrinf_case}
 \end{align}
Thus, as a function of measurement density, the high SNR behavior of quadratic inference exhibits a phase transition at the critical density $\alpha_c=1$.  Below this density, in the undersampled regime, performance asymptotes to a finite error, independent of SNR. Above this density, in the oversampled regime, inference error decays with SNR as $\snr^{-1}$.  Surprisingly, at the critical density, the decay with SNR is slower, and exhibits a universal decay exponent of $-\frac{1}{2}$, independent of the signal and noise distributions.  This exponent, and its universality, is verified numerically in Fig. \ref{quadplot}A.   Moreover, as $\alpha \rightarrow 1$, $\bar q^{\text{Quad}}_s$, remains $O(1)$ at any finite $\snr$, unlike the unregularized case.  Indeed, for $\alpha \ll 1$,  $\bar q^{\text{Quad}}_s =  1 - \alpha \frac{\snr}{\snr + 1}$. Thus quadratic regularization can tame the divergence of unregularized inference at low measurement density.
\par
The phase transition behavior of optimal quadratic inference can be understood from the perspective of random matrix theory (RMT).  In the special case of \eqref{mEstEq} when $\rho(x) = \frac{1}{2}x^2$ and $\sigma(x) = \frac{1}{2} \frac{1}{\text{SNR}} x^2$, the optimal estimate $\hat \s$ has the analytic solution
\begin{equation}
\hat{\s} = \left(\mathbf{X}^T \mathbf{X} + \frac{1}{\text{SNR}} \mathbf{I}\right)^{-1}\mathbf{X}^T \mathbf{y},
\end{equation}
where $\mathbf{X}$ is an $N$ by $P$ measurement matrix whose $N$ rows are the $N$ measurement vectors $\x^\mu$ (see \cite{PRLsupp} Sec. 3.5 for more details).  This analytic solution for $\hat \s$ enables a direct average over the noise $\boldsymbol{\epsilon}$ and true signal $\s^0$ in $\mathbf{y}$ to yield
\begin{equation}
\bar q^{\text{Quad}}_s =  \frac{1}{P} \text{Tr}  \left [ \mathbf{I} + \text {SNR} \, \mathbf{X}^T \mathbf{X} \right]^{-1}.
\label{eq:qsbeigval1}
\end{equation}
This expression can be reduced to an average over the eigenvalue distribution of the random measurement correlation matrix $\mathbf{X}^T \mathbf{X}$, which has the well known Marcenko-Pasteur (MP) form \cite{marchenko1967distribution}:
\begin{equation}
\pmp(\lambda) = \frac{1}{2\pi} \frac{\sqrt{(\lambda_+ - \lambda)(\lambda - \lambda_-)}}{\lambda} + \mathbf{1}_{\alpha<1} (1 - \alpha)\delta(\lambda),
\label{eq:marpas}
\end{equation}
where the nonzero support of the density is restricted to the range $\lambda \in [\lambda_-, \lambda_+]$, with $\lambda_{\pm} = \left(\sqrt{\alpha} \pm 1\right)^2$.  Also $\mathbf{1}_{\alpha<1}$ is $1$ when $\alpha < 1$ and $0$ otherwise. Thus at measurement densities $\alpha < 1$, the MP distribution has an additional delta function at the origin with weight $1-\alpha$, reflecting the fact that the $P \times P$ measurement correlation matrix $\mathbf{X}^T \mathbf{X}$ is not full rank when $N < P$.  In terms of $\pmp(\lambda)$, \eqref{eq:qsbeigval1} reduces to
\begin{equation}
\bar q^{\text{Quad}}_s = \int{  \Delta(\lambda) \, \pmp(\lambda) \, d\lambda},
\label{eq:qsbintegral}
\end{equation}
where $\Delta(\lambda) = (1 +\lambda \cdot \snr )^{-1}$.  Direct calculation reveals that expression \eqref{eq:qsbintegral} for $\bar q^{\text{Quad}}_s(\alpha, \text{SNR})$, derived via random matrix theory, is consistent with the expression \eqref{eq:qsquadopt}, derived via our theory of high dimensional statistical inference.
\par
The expression for $\bar q^{\text{Quad}}_s$ in \eqref{eq:qsbintegral} can now be used to elucidate the nature of the phase transition in Fig. \ref{quadplot}A.  At high SNR, the function $\Delta(\lambda)$ remains $O(1)$ in a narrow regime of width $O(\frac{1}{\text{SNR}})$ near the origin.  However, when $\alpha < 1$, the left edge $\lambda_-$ of the nonzero part of the MP density remains separated from the origin.  Due to this eigenvalue density gap, the dominant contribution to the integral in \eqref{eq:qsbintegral} arises from the $\delta$-function at the origin, yielding $\bar q^{\text{Quad}}_s \approx 1-\alpha$ when $\alpha < 1$ (see Fig. \ref{quadplot}B top).  When $\alpha > 1$, the $\delta$-function is absent and the dominant contribution arises from the nonzero part of the MP density.  This density has support over a range that is $O(\alpha)$ yielding $\bar q^{\text{Quad}}_s =
O(\frac{1}{\text{SNR} \, \alpha})$ (see Fig. \ref{quadplot}B bottom).  Only when $\alpha=1$ does the gap in the MP density vanish.  In this case, near the origin, the density
diverges as $\lambda^{-1/2}$ (see Fig. \ref{quadplot}B middle).  At high SNR, because $\Delta(\lambda)$ induces an effective cut-off at $\frac{1}{\text{SNR}}$, the integral in \eqref{eq:qsbintegral} can be approximated as $\int_{0}^{\text{SNR}^{-1}} \lambda^{-1/2} \, \text{d}\lambda = O(\text{SNR}^{-1/2})$.
\par Thus the origin of the phase transition in \eqref{snrinf_case} at the critical value $\alpha=1$ arises from the vanishing of a gap in the MP distribution.  Moreover, the universal decay exponent at the critical value of $\alpha=1$ is related to the power law behavior of the MP density near the origin at $\alpha=1$.  Remarkably, this highly nontrivial behavior is captured simply through the outcome of our replica analysis for optimal quadratic inference, encapsulated in the pair of equations in \eqref{eq:qdqsquad}.

\subsubsection{The worst signal and noise distributions are Gaussian}
\par We note that this optimal quadratic inference procedure is optimal amongst all possible inference procedures, if and only if the signal and noise are Gaussian, since, in that case, it is equivalent to the Bayesian MMSE inference procedure.  Moreover, we note that Gaussian signal and noise are in some sense the {\it worst} type of signal and noise distributions, in the space of all inference problems with a given SNR.  To see this, consider a non-Gaussian signal and noise with a given SNR.  The performance of optimal quadratic inference for this non-Gaussian signal and noise only depends on the pair of distributions through their SNR, and is equivalent to the performance of optimal quadratic inference for Gaussian signal and noise at the same SNR.  However, in the non-Gaussian case, a non-quadratic inference algorithm could potentially outperform the quadratic one, but not in the Gaussian case, since quadratic inference is already optimal in that case.  Thus in the space of inference problems of a given SNR, the worst case performance of optimal inference occurs when both the signal and noise are Gaussian.

\subsubsection{Optimal inference with non-Gaussian signal and noise}
\par
What is the optimal (non-quadratic) inference procedure in the face of non-Gaussian signal and noise?  We address this by performing a functional minimization of $q_s$ over {\it both} $\rho$ and $\sigma$, subject to constraints \eqref{rsreg1} and \eqref{rsreg2}, which yields (\cite{PRLsupp}, sec. 5.2),
%
\begin{align}
 \rhoopt(x) &=  -\mor{\qsopt}{-E_{\epsilon_{\qsopt}}}{x} ,\label{eq:optRhoReg} \\
\sigopt(x)  & =-\mor{\qoptrho}{-E_{s_{\qoptrho}}}{x}, \label{eq:optSigReg}
\end{align}
where $\qsopt$ and $\qoptrho$ satisfy
\begin{equation}
\qoptrho   = \frac{1}{\alpha\fish{\epsilon_{\qsopt}}}, \quad
\qsopt  = q_s^\text{MMSE}(\qoptrho),
\label{eq:regopt}
\end{equation}
and the function $q_s^\text{MMSE}$ is defined in \eqref{eq:qmmse}.  Again, the validity of \eqref{eq:optRhoReg}-\eqref{eq:optSigReg} under the RS assumption requires convexity of $\rhoopt$ and $\sigopt$.  Convexity of the signal and noise energies, $E_s$ and $E_\epsilon$ are sufficient to guarantee convexity of $\rhoopt$ and $\sigopt$, and so for this class of signal and noise, with log concave distributions, \eqref{eq:optRhoReg}-\eqref{eq:optSigReg} yields an optimal inference procedure.  However, by judicious applications of the Cauchy-Schwarz inequality, we prove (\cite{PRLsupp}, sec. 4.1) that even for non-convex $E_s$ and $E_\epsilon$, the inference error $q_s$ for ${\it any}$ convex procedure $(\rho,\sigma)$ must exceed $\qsopt$ in \eqref{eq:regopt}.  This result yields a fundamental limit on the performance of any convex inference procedure of the form \eqref{mEstEq} in high dimensions.

\par Intriguingly, by comparing the optimal achievable high dimensional M-estimation performance $\qsopt$ in \eqref{eq:regopt} to the asymptotic performance of low dimensional scalar Bayesian inference in \eqref{eq:qdnoiseeff} and \eqref{eq:qmmse}, we find a striking parallel.  In particular, $\qsopt$ corresponds to the low dimensional asymptotic MMSE in a scalar estimation problem where the effective number of measurements $N = \alpha$ and the noise $\epsilon$ is further corrupted by additional Gaussian noise of variance $\qsopt$ ($\epsilon \rightarrow  \epsilon + \sqrt{\qsopt} z$).  The correction to the low dimensional scalar asymptotics \eqref{eq:qmmse}, valid only at large $N$, in the high dimensional regime at finite measurement density $\alpha$, is obtained by self-consistently solving for $\qsopt$ in \eqref{eq:regopt}.  In essence, at finite measurement density, there is irreducible error in estimating the signal, $\qsopt$.  This error contributes to the effective Gaussian noise $\qoptrho$ in the scalar MFT estimation problem for the signal, shown in Fig. \ref{fig:schema}B, where the proximal map becomes the Bayesian posterior mean map in the optimal case.  On the otherhand, this irreducible, extra gaussian noise is absent in low dimensions (compare LHS of \eqref{eq:regopt} to {\eqref{eq:qdnoiseeff}).  This irreducible error $\qsopt$ can be found by self-consistently solving for it in the RHS of \eqref{eq:regopt}.  Finally, as a simple point, we note that direct calculation reveals that \eqref{eq:regopt} reduces to \eqref{eq:qdqsquad}} when the signal and noise are both Gaussian distributed, as expected, since optimal quadratic inference is the best procedure for Gaussian signal and noise.

\par Furthermore, using the fact that the equalities in \eqref{eq:regopt} become inequalities for non-optimal procedures (\cite{PRLsupp}, section 4.2), we can derive a high dimensional analogue of \eqref{eq:bayesbound}, and prove a lower bound on the inference error $q_s$ for any convex $(\rho, \sigma)$:
\begin{equation}
q_s \geq \frac{1}{\alpha \fish{\epsilon_{q_s}} + \fish{s^0}}.
\label{eq:highbayesbound}
\end{equation}
This results reflects a fundamental generalization of the high-dimensional CR bound \eqref{optConNB} that includes information about the signal distribution $P_s$ that can be optimally exploited by a regularizer $\sigma$.  Since $\fish{\epsilon_{\qs}} < \fish{\epsilon}$, by the data processing inequality for Fisher information, this high dimensional lower bound is larger than the low-dimensional one \eqref{eq:bayesbound} under the replacement $\alpha \rightarrow N$.  Thus, as in the unregularized case \eqref{optConNB}, the price paid for even optimal high-dimensional regularized inference at finite measurement density, relative to scalar Bayesian inference at asymptotically infinite density, is increased error due to the presence of additional gaussian noise with dimensionality dependent variance $\qsopt$.

\subsubsection{Optimal high dimensional inference smoothly interpolates between MAP and quadratic inference}

\par The optimal inference procedure \eqref{eq:optRhoReg}-\eqref{eq:optSigReg} is a smoothed version of MAP inference (see Fig. \ref{q0plot}C for an example of smoothing), where the MAP choices $\rho = E_\epsilon$ and $\sigma = E_s$  are smoothed over scales $\qsopt$ and $\qoptrho$ respectively to obtain $\rhoopt$ and $\sigopt$.  As $\alpha \rightarrow \infty$, both $\qsopt$ and $\qoptrho$
approach $0$ at the same rate, implying $\rhoopt \rightarrow E_\epsilon$ and $\sigopt \rightarrow E_s$.  Thus at high measurement density, MAP inference is the optimal M-estimator.  This conclusion is intuitively reasonable because at high measurement densities, the mode of the posterior distribution over the signal, returned by the MAP estimate, is typically close to the mean of the posterior distribution, which is the optimal MMSE estimate amongst all inference procedures.
\par
\begin{figure*}[t]
\begin{center}
\includegraphics[width=0.65\textwidth]{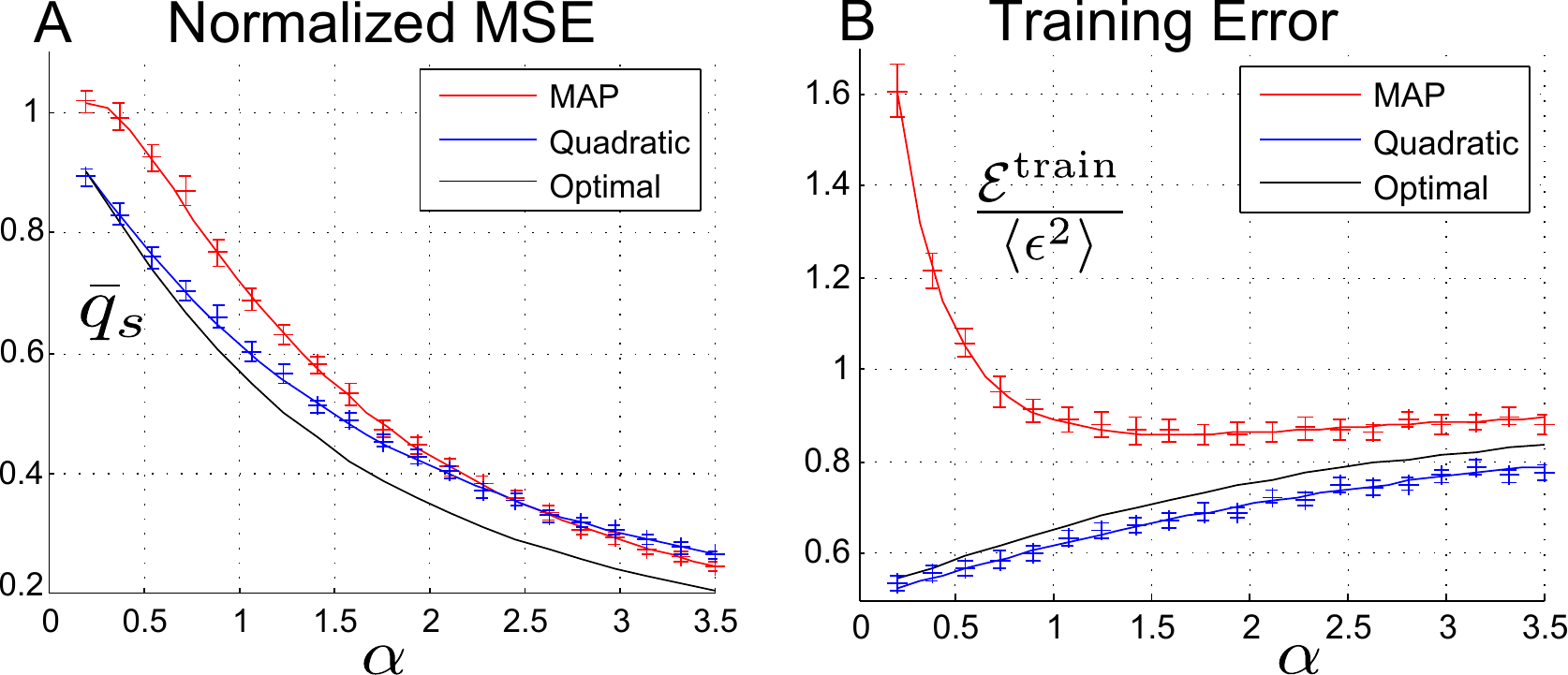}
\caption{Regularized inference for Laplacian noise and signal $E_{\epsilon} = |\epsilon|, E_s =|s^0|$. (A) The normalized MSE, or fraction of unexplained variance $\bar q_s$.  (B) The training error.  Each plot shows the respective performance of $3$ different inference procedures: our optimal inference (\ref{eq:optRhoReg},\ref{eq:optSigReg}) (black), MAP inference (red), and optimal quadratic inference (blue). The theoretical predictions (solid curves) match numerical simulations (error bars) which reflect the standard deviation calculated over 20 trials using a convex optimization solver for randomly generated, finite sized data (with $N$ and $P$ varying while $N = \alpha P$ and $\sqrt{N P} = 250$). Note that optimal inference can significantly outperform common but suboptimal methods. For example to achieve a fraction of unexplained variance of $0.4$, optimal inference requires a measurement density of $\alpha \approx 1.7$ while quadratic and MAP inference require $\alpha \approx 2.1$ and $\alpha \approx 2.2$ respectively. This reflects a reduction of approximately $20$ percent in the amount of required data.}
\label{RegCompare}
\end{center}
\end{figure*}

  Alternatively, as $\alpha \rightarrow 0$, $\qsopt \rightarrow \langle {s^2} \rangle$ from below, while $\qoptrho$ diverges as $\frac{1}{\alpha}$.   The divergence of $\qoptrho$ implies that $\sigma_{opt}$ in \eqref{eq:optSigReg} approaches a quadratic. Thus, remarkably, at low measurement density, simple quadratic regularization, independent of the signal distribution, becomes an optimal inference procedure.  Furthermore, in the low density plus high SNR limit, where $\langle \epsilon^2 \rangle \ll \langle s^2 \rangle$,  $\rhoopt$ also approaches a quadratic.  Thus overall, optimal high dimensional inference at high SNR interpolates between MAP and quadratic inference as the measurement density decreases.  In Figure \ref{RegCompare} we demonstrate, for Laplacian signal and noise,  that optimal inference outperforms both MAP and quadratic inference at all $\alpha$, approaching the former at large $\alpha$ and the latter at small $\alpha$.

\subsubsection{A relation between optimal high dimensional inference of signal, and low-dimensional Bayesian inference of noise}
\par
There is an interesting connection between optimal high dimensional inference, and low-dimensional scalar Bayesian inference.  Indeed, when $\rho$ and $\sigma$ take their optimal forms in \eqref{eq:optRhoReg} and \eqref{eq:optSigReg}, then the proximal descent steps in \eqref{eq:rsprox} used to estimate noise and signal in the pair of coupled estimation problems comprising the MFT (shown schematically in Fig. \ref{fig:schema}AB) become optimal Bayesian estimators. In particular, for optimal $\rho$ and $\sigma$, \eqref{eq:rsprox} becomes (\cite{PRLsupp}, section 5.2)

\begin{equation}
\hat \epsilon(\eqs) =  \langle \, \epsilon \, | \, \epsilon_{q_s} \, \rangle   \qquad  \hat s(\sqr) = \langle \, s \, | \, \sqr \, \rangle.
\label{eq:scalarpostmean}
\end{equation}
In essence, computation of the proximal map becomes computation of the posterior mean, which is the optimal, MMSE method for estimating signal and noise in the MFT scalar estimation problems.  This gives an intuitive explanation for the form of $\rhoopt$ and $\sigopt$ in \eqref{eq:optRhoReg} and \eqref{eq:optSigReg}: these are exactly the forms of loss and regularization required for the proximal descent estimates in \eqref{eq:rsprox}  to become optimal posterior mean estimates in \eqref{eq:scalarpostmean}.

\subsubsection{A relation between signal-noise separation, and predictive power}

\par Furthermore, there is an interesting connection between our ability to optimally estimate noise and signal, and the training and test error.  In particular, just as our error $\qsopt$ in estimating the signal is given by \eqref{eq:regopt} and \eqref{eq:qmmse}, our error in estimating the noise is given by $\qeopt = \qav{ (\hat \epsilon - \epsilon)^2 }$,  with $\hat \epsilon$ given in \eqref{eq:scalarpostmean}, yielding
\begin{equation}
\qeopt = q_\epsilon^\text{MMSE}(\qsopt) = \qav{ \left(\epsilon - \langle \epsilon  \, | \, \epsilon_{\qsopt} \rangle \right)^2 }.
\end{equation}
In terms of these quantities, the generalization and training errors of the optimal M-estimator have very simple forms (\cite{PRLsupp}, section 5.2):
\begin{equation}
\Etrain  = \langle \epsilon^2 \rangle  -  \qeopt  \qquad  \Egen = \langle \epsilon^2 \rangle + \qsopt.
\label{eq:gentrainbayes}
\end{equation}
This leads to an intuitively appealing result: inability to estimate the signal leads directly to increased generalization error, while inability to estimate the noise leads to {\it decreased} training error.
\par The reason for this latter effect is that if the optimal inference procedure cannot accurately separate signal from noise to correctly estimate the noise, then it mistakenly identifies noise in the training data as signal, and this noise is incorporated into the parameter estimate $\hat \s$.   Thus $\hat \s$ acquires correlations with the particular realization of noise in the training set so as to reduce training error. However,  this reduced training error comes at the expense of increased generalization error, due again to mistaking noise for signal.  The predicted decrease of training error and increase of generalization error  for the optimal inference procedure as measurement density decreases is demonstrated in Fig. \ref{RegCompare}.   Interestingly, this figure also demonstrates that training error need not decrease at low measurement density for suboptimal algorithms, like MAP.
\par Thus, in summary, the ability to correctly separate signal from noise to extract a model of the measurements $\mathbf{y}$ in \eqref{mEstEq} is intimately related to the predictive power of the extracted model $\hat \s$ in \eqref{mEstEq}.  Inability to estimate noise reduces training error, while inability to estimate signal increases generalization error.  The combination is a hallmark of overfitting the learned model parameters to the training data, and thereby incurring a loss of predictive power on new, held-out data.

\subsection{Inference without noise}

Motivated by compressed sensing, there has been a great deal of interest in understanding when and how we can perfectly infer the signal, so that $q_s=0$, in the undersampled measurement regime $\alpha < 1$.  This can only be done in the absence of noise ($\epsilon = 0$), but what properties must the signal distribution satisfy to guarantee such remarkable performance?  In this special case of no noise, $\epsilon_{q_s}$ simply becomes a Gaussian variable with variance $q_s$, with Fisher information $J[\epsilon_{q_s}] = \frac{1}{q_s}$.   Using this, and a relation between MMSE and Fisher information (\cite{PRLsupp}, appendix B.4),  the optimality equations in \eqref{eq:regopt} become
 \begin{equation}
\qoptrho   = \frac{\qsopt}{\alpha} \qquad
\qsopt  = \qoptrho \left( 1- \qoptrho J[s^0_{\qoptrho}] \right).
\label{eq:regoptnoiseless}
\end{equation}
Partially eliminating $\qoptrho$ yields
\begin{equation}
\qsopt = \frac{\alpha(1-\alpha)}{J[s^0_{\qoptrho}]} \geq \frac{1-\alpha}{J[s^0]}.
\label{eq:noiselessbound}
\end{equation}
Here the inequality arises through an application of the convolutional Fisher inequality
\begin{equation}
\frac{1}{J[s^0_{\qoptrho}]} \geq \frac{1}{J[s^0]} + \qoptrho,
\end{equation}
and then fully eliminating $\qoptrho$.
\par Given that for any signal and noise distribution, we have proven that no convex inference procedure can achieve an error smaller than $\qsopt$, \eqref{eq:noiselessbound} yields a general, sufficient, information theoretic condition for perfect recovery of the signal in the noiseless undersampled regime: the Fisher information of the signal distribution must diverge.  This condition holds for example in sparse signal distributions that place finite probability mass at the origin.  More generally, \eqref{eq:noiselessbound} yields a simple lower bound on noiseless, undersampled inference in terms of the measurement density and signal Fisher information.  Moreover, in situations where the signal energy is convex, \eqref{eq:optSigReg} remains the optimal inference procedure, while $\rhoopt$ is replaced with a hard constraint enforcing optimization only over candidate signals $\s$ satisfying the noiseless measurement constraints $y^\mu = \x^\mu \cdot \hat{s}$.


\section{Discussion}
In summary, our theoretical analyses, verified by simulations, yield a fundamental extension of time honored results in low-dimensional classical statistics to the modern regime of high dimensional inference, relevant in the current age of big data.
In particular, we characterize the performance of any possible convex inference procedure for arbitrary signal and noise distributions (Eqs. \ref{rsreg1}-\ref{rsreg2}), we find fundamental information theoretic lower bounds on the error achievable by any convex procedure for arbitrary signal and noise (Eq. \ref{eq:highbayesbound}), and, we find the inference procedure that optimally exploits information about the signal and noise distributions, when their energies are convex (Eqs. \ref{eq:optRhoReg}-\ref{eq:optSigReg}).  Moreover we find a simple information theoretic condition for successful compressed sensing (Eq. \ref{eq:noiselessbound}), or perfect inference without full measurement.  These results generalize classical statistical results, based on Fisher information and the Cramer-Rao bound, that were discovered over $60$ years ago.

\par Moreover, our analysis uncovers several interesting surprises about the nature of optimal high dimensional inference.  In particular, we find that the optimal high dimensional inference procedure is a smoothed version of ML in the unregularized case, and a smoothed version of MAP in the regularized case, where the amount of smoothing increases as the measurement density decreases, or equivalently as the dimensionality increases.  At low measurement densities and high dimensions, the optimal smoothed loss and regularization functions become simple quadratics (in the regularized case, this is proveably true strictly at high SNR, but empirically, replacing the optimal loss with quadratic loss incurs very little performance decrement even at moderate SNR (Fig. \ref{RegCompare}A)).  This observation reveals a fortuitous interplay between problem difficulty and algorithmic simplicity: at low measurement density, precisely when inference becomes statistically difficult, the optimal algorithm becomes computationally simple.  Finally, we uncover phase transitions in the behavior of this simple quadratic inference algorithm, with a universal critical exponent in the decay of inference error with SNR at a critical measurement density (Eq. \ref{snrinf_case}).

\par  Also, our analyses reveal several conceptual insights into the nature of overfitting and generalization in optimal high dimensional inference through novel connections scalar to Bayesian inference in one dimension.    This connection arises due to the nature of the mean field theory of general high dimensional inference, which can be expressed in terms of two coupled scalar estimation problems for the noise and signal respectively (Fig. \ref{fig:schema}).   In the optimal case, these scalar inference procedures based on proximal descent steps (Eq. \ref{eq:rsprox}) become Bayesian inference procedures (Eq. \ref{eq:scalarpostmean}).  In particular, any inference algorithm implicitly decomposes the given measurements $y^\mu = \x^\mu \cdot \s^0 + \epsilon^\mu$ into a superposition of {\it estimated} signal and {\it estimated} noise: $y^\mu = \x^\mu \cdot \hat \s  + \hat \epsilon^\mu$.   The scalar Bayesian inference problems yield a MFT prediction for the error in estimating the signal (average per component $L_2$ discrepancy between $\s$ and $\hat \s$) and noise (average per component $L_2$ discrepancy between $\epsilon^\mu$ and $\hat\epsilon^\mu$).   Errors in inference arise because the noise $\epsilon^\mu$ seeps into the estimated signal $\hat \s$.  This inability to accurately separate signal and noise by even the optimal inference algorithm leads to divergent effects on the training and generalization error. The former decreases as the estimated signal $\hat \s$ acquires spurious correlations with the true noise $\epsilon^\mu$ to explain the measurement outcomes $y^\mu$. The latter increases because the noise in a held out, previously unseen measurement outcome cannot possibly be correlated with the signal $\hat \s$ estimated from previously seen training data.  Indeed, for the optimal inference algorithm, we find exceedingly simple quantitative relationships between inference errors of noise and signal, and high dimensional training and generalization error (Eq. \ref{eq:gentrainbayes}).  This yields both quantitative and conceptual insight into the nature of overfitting in high dimensions, whereby training error can be far less than generalization error.

Overall, our results illustrate the power of statistical mechanics based methods to generalize classical statistics to the new regime of high dimensional data analysis.  We hope that these results will provide both firm theoretical guidance, as well as practical algorithmic advantages in terms of both statistical and computational efficiency,  to many fields spanning the ranges of science, engineering and the humanities, as they all attempt to navigate the brave new-world of big-data.







\section*{Acknowledgements}

We thank Subhaneil Lahiri for useful discussions and also Alex Williams and Niru Maheswaranathan for comments on the manuscript. M.A. thanks the Stanford MBC and SGF for support.  S.G. thanks the Burroughs Wellcome, Simons, Sloan, McKnight, and McDonnell foundations for support.

\bibliography{PRLcite_edited,suryabib10}{}
\bibliographystyle{unsrt}

\end{document}